\documentclass[10pt,a4paper,twocolumn]{article}

\usepackage{graphicx} 

\usepackage{subfigure} 
\usepackage{subfigmat} 

\usepackage{dcolumn}           
\newcolumntype{.}   {D{.}{.}{-1}} 
\newcolumntype{d}[1]{D{.}{.}{#1}} 
\newcolumntype{e}   {D{E}{E}{-1}} 
\newcolumntype{E}[1]{D{E}{E}{#1}} 

\usepackage{amsmath}
\usepackage{mathptmx}

\usepackage{amssymb}
\usepackage{amsthm}

\usepackage{float}

\usepackage{setspace}

\usepackage{sectsty}

\newcommand{\myFontSize}{\fontsize{10}{12}\selectfont}
\renewcommand{\normalsize}{\fontsize{10}{12}\selectfont}
\sectionfont{\myFontSize}       
\subsectionfont{\rm\myFontSize\itshape} 

\usepackage{titlesec}
\titlespacing*{\section}{0pt}{10pt}{0pt}
\titlespacing*{\subsection}{0pt}{10pt}{0pt}

\usepackage{secdot}
\sectiondot{subsection}
\sectionpunct{section}{. }    
\sectionpunct{subsection}{. } 

\usepackage{caption}
\usepackage{xcolor} 
\usepackage{fancyhdr} 
\usepackage{lastpage}
\renewcommand{\headrulewidth}{0pt} 

\title{\textbf{\selectfont\fontsize{10}{0}\selectfont Taxonomy for Resident Space Objects in LEO}}
\date{}
\author{
    \textbf{
        \selectfont\fontsize{10}{0}\selectfont Marta~Guimarães~\textsuperscript{a*},~Cláudia~Soares~\textsuperscript{b},~Chiara Manfletti\textsuperscript{a} 
    } \\ 
}

\usepackage[numbers]{natbib}

\usepackage[shortcuts,acronym,nonumberlist]{glossaries}
\makeglossaries 
\newacronym{LEO}{LEO}{low-Earth orbit}
\newacronym{RSO}{RSO}{resident space object}
\newacronym{RCS}{RCS}{radar cross-section}
\newacronym{DISCOS}{DISCOS}{Database and Information System Characterising Objects in Space}
\newacronym{ESA}{ESA}{European Space Agency}
\newacronym{SATCAT}{SATCAT}{Satellite Catalog}

\newacronym{UMAP}{UMAP}{Uniform Manifold Approximation and Projection}
\newacronym{DL}{DL}{Deep Learning}
\newacronym{SSE}{SSE}{sum of squared error}
\newacronym{SHAP}{SHAP}{Shapley additive explanations}

\begin{document}

%
\twocolumn[
\begin{@twocolumnfalse}

\vspace{0pt}
\begin{center}
    
    \vspace{15pt}
    \textbf{\selectfont\fontsize{10}{0}\selectfont Taxonomy for Resident Space Objects in LEO: A Deep Learning Approach}
    
    \vspace{10pt}
    \textbf{
        \selectfont\fontsize{10}{0}\selectfont Marta~Guimarães~\textsuperscript{a,b*},~Cláudia~Soares~\textsuperscript{b},~Chiara Manfletti\textsuperscript{a} 
    }
\end{center}


\vspace{-10pt} 
\begin{flushleft}
    \textsuperscript{a}\textit{
    \fontfamily{ptm}\selectfont\fontsize{10}{12}\selectfont Neuraspace, Portugal}, \underline{\{marta.guimaraes, chiara.manfletti\}@neuraspace.com}
    \\
    \textsuperscript{b}\textit{
        \fontfamily{ptm}\selectfont\fontsize{10}{12}\selectfont FCT-UNL, Portugal},
        \underline{claudia.soares@fct.unl.pt}
    \\
    \textsuperscript{*}\fontfamily{ptm}\selectfont\fontsize{10}{12}\selectfont Corresponding Author  
\end{flushleft}

\begin{abstract}
The increasing number of \glspl{RSO} has raised concerns about the risk of collisions and catastrophic incidents for all direct and indirect users of space. To mitigate this issue, it is essential to have a good understanding of the various \glspl{RSO} in orbit and their behaviour. A well-established taxonomy defining several classes of \glspl{RSO} is a critical step in achieving this understanding. This taxonomy helps assign objects to specific categories based on their main characteristics, leading to better tracking services. Furthermore, a well-established taxonomy can facilitate research and analysis processes by providing a common language and framework for better understanding the factors that influence \gls{RSO} behaviour in space. These factors, in turn, help design more efficient and effective strategies for space traffic management. Our work proposes a new taxonomy for \glspl{RSO} focusing on the low Earth orbit regime to enhance space traffic management. In addition, we present a deep learning-based model that uses an autoencoder architecture to reduce the features representing the characteristics of the \glspl{RSO}. The autoencoder generates a lower-dimensional space representation that is then explored using techniques such as Uniform Manifold Approximation and Projection to identify fundamental clusters of \glspl{RSO} based on their unique characteristics. This approach captures the complex and non-linear relationships between the features and the \glspl{RSO}’ classes identified. Our proposed taxonomy and model offer a significant contribution to the ongoing efforts to mitigate the overall risks posed by the increasing number of \glspl{RSO} in orbit.


\end{abstract}

\end{@twocolumnfalse}
]


\section{Introduction}

\subsection{Space Safety and Sustainability}
The increasing number of \glspl{RSO} in orbit, such as satellites and debris, has caused concerns about potential collisions and catastrophic incidents for both direct and indirect users of space~\cite{Kessler1978, Virgili2016}. As the number of satellites being launched continues to increase and the amount of debris proliferates, it is becoming more difficult to manage space traffic and mitigate the risks associated with \glspl{RSO}~\cite{Virgili2019}.

Moreover, this issue is not limited to the realm of space exploration and technology but also extends to the fields of international relations and national security. As space becomes more crowded with \glspl{RSO}, the risk of collisions and incidents increases~\cite{ESA2022}, which could have severe consequences for countries that rely on satellites for communication, navigation, and national defence.

To ensure the safety and sustainability of space activities for generations to come, governments and space agencies must work together to address the concerns surrounding \glspl{RSO}~\cite{UNOOSA2022}.
A well-established taxonomy that defines the several classes of \glspl{RSO} is critical in better understanding these objects and their behaviour in space. Such a taxonomy can help assign objects to specific categories based on their main characteristics, leading to better tracking services. As the \gls{LEO} becomes more crowded with heterogeneous objects the interactions between such objects become intractable by human eye inspection, even for highly trained experts.
Furthermore, a well-established taxonomy can facilitate research and analysis processes by providing a common language and framework for better understanding the factors that influence \gls{RSO} behaviour. These factors, in turn, help to improve strategies for space traffic management. Therefore, a new data-driven taxonomy for \glspl{RSO} focusing on the \gls{LEO} regime offers a significant contribution to the ongoing efforts to mitigate the overall risks posed by the increasing number of \glspl{RSO} in orbit.

In this work, we propose a new taxonomy for \glspl{RSO} focusing on the \gls{LEO} regime. Our taxonomy is based on an extensive analysis of the different types of \glspl{RSO} and their characteristics in order to provide a more comprehensive understanding of the space debris environment. We also present a \gls{DL}-based model that uses an autoencoder architecture to distil the features representing the characteristics of the \glspl{RSO} into the most significant information. This model is trained on a large dataset of \glspl{RSO} and is capable of generating a lower-dimensional space representation that can be used to identify fundamental clusters of \glspl{RSO} based on their unique characteristics.

Our proposed taxonomy and \gls{DL}-based representation model constitute an important step forward in the field of space debris management, providing new insights into the complex dynamics of the \gls{LEO} regime and enabling more effective mitigation strategies to protect critical space infrastructure.

\subsection{Related Work}
Due to the rapidly increasing number of \glspl{RSO}, there has been a growing interest in establishing a unified taxonomy to classify them. This is because a common taxonomy facilitates the comparison of different \glspl{RSO} and their characteristics, allowing for a better understanding and organisation of the field. Several works in the literature have proposed different taxonomies for \glspl{RSO}, each with its own strengths and weaknesses. While there is no consensus on the ideal taxonomy, ongoing research and discussion in the field aim to develop a unified framework that can capture the full scope of \glspl{RSO} and their diverse characteristics.

~\citet{Fruh2013} was the first to introduce a more sophisticated taxonomy for \glspl{RSO} inspired by taxonomy schemes used in biology. The proposed solution established a hierarchical tree based on prior empirical knowledge. Then, hierarchical clustering based on the \gls{RCS} was performed to analyse the proposed taxonomy.~\citet{Jah2013} leveraged the previous work and proposed a new \gls{RSO} taxonomy along with its implementation using probabilistic programming. Later, this work was expanded by~\citet{Wilkins2014} to explore the implications of \gls{RSO} taxonomic trees for model distance metrics and sensor tasking. Recent work of~\citet{Mellish2021} developed a fully-automated taxonomy of geosynchronous objects based on dynamical principles.

While taking inspiration from previous work, our approach focuses on the \gls{LEO} regime and is derived from the data.

\subsection{Contributions}
We propose a novel taxonomy for \glspl{RSO} in \gls{LEO} derived by analysing a large dataset of catalogued objects. Our approach combines both real-valued and categorical variables and considers missing values, avoiding the imputation of default data. By considering missing values as valuable information, we ensure that the taxonomy reflects the true nature and characteristics of the objects, without making assumptions or artificially filling in data, providing a more robust and unbiased model. Furthermore, our taxonomy organises the objects into distinct categories based on their intrinsic characteristics and orbital parameters, allowing for a more accurate representation of the underlying patterns and relationships of \glspl{RSO} in \gls{LEO}.

We acknowledge and discuss the limitations of current satellite datasets in providing comprehensive information such as the level of automation of \glspl{RSO} or classification criteria for emerging satellite shapes.

\section{Data Sources} \label{sec:data}

In this study, we used a dataset of catalogued space objects obtained from the publicly available \gls{SATCAT}\footnote{https://celestrak.org/satcat/search.php}. Such a dataset consists of 55.5k objects~\footnote{Corresponding to the number of objects registered at the beginning of 2023.} and contains mainly information about the orbital characteristics of the objects such as perigee, apogee and inclination. The \gls{DISCOS}\footnote{https://discosweb.esoc.esa.int/}~\cite{Flohrer2013} was also used to complement this information. The new features introduced include information about the main characteristics of the objects, e.g., size, mass, and shape.

After merging these data, only the samples concerning objects in \gls{LEO} were kept resulting in a dataset with 20.5k samples. The resulting dataset contains both real-valued and categorical features, providing a diverse range of information about each object. Throughout this work, the missing values were handled with appropriate techniques, ensuring that we were able to make use of as much available information as possible.

\section{Autoencoder Representation Model}

As mentioned in Section \ref{sec:data}, the input dataset contains both real-valued and categorical features. In order to use such data to train a \gls{DL} model, it is necessary first to process the data to ensure that it is in a suitable format. 

\begin{figure*}[t!]
    \centering
    \includegraphics[width=\textwidth]{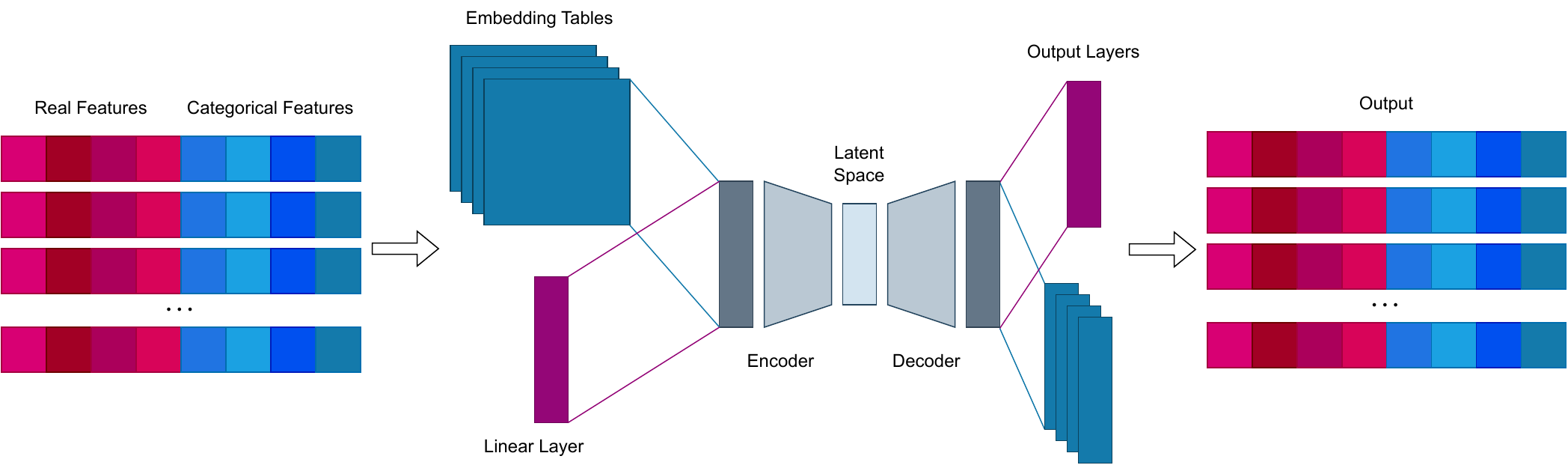}
    \caption{Architecture of the autoencoder with real and categorical features. The real features (shades of red/pink) are processed through a linear layer, while the categorical features (presented in blue) are transformed using embedding tables to ensure that all features are represented in a common embedding space. The resulting features are added into a single vector (presented in dark grey) and passed through an autoencoder to compress the data (in light grey), with the output reconstructed back to the original dimension through additional linear layers (labelled as output layers).}
    \label{fig:autoencoder}
\end{figure*}

For the real-valued features, we normalize them to have zero mean and unit variance, which is a common preprocessing step for continuous data that have different ranges of values. Then, these features are mapped to a different dimension space through a linear layer (Figure~\ref{fig:autoencoder}). This step is needed to ensure that these features can be combined with the remaining ones.

On the other hand, the categorical features are discrete and represented mainly by strings and, thus, cannot be directly used in a \gls{DL} model. Therefore, we convert them into a continuous representation using embedding tables~\cite{Mikolov2013}.
Embedding tables are a commonly used technique in natural language processing tasks. The basic concept underlying embedding tables is to represent categorical variables as low-dimensional vectors, where each dimension of the vector corresponds to a particular category. The embeddings are learned during the training process, using a loss function that encourages similar categories to be mapped to nearby points in the embedding space. A common loss function to be minimised during training for this problem is the categorical cross-entropy loss, which measures the difference between the predicted and the true categories.

Once the real-valued and categorical features have been mapped to the same dimension in Euclidean space, the outputs of these layers are then fused to create a single feature representation. Such fusion is done by summing all the vectors.
This resulting feature vector is then passed through an autoencoder~\cite{Zhai2018}.

The autoencoder architecture consists of two primary components: an encoder and a decoder. The encoder takes in the input data and maps it to the latent space, while the decoder takes the compressed latent representation and maps it back to the original input space. The network is trained with the aim of minimising the reconstruction error between the input and the output. By minimising this reconstruction error, the autoencoder is able to learn a compressed representation of the input data that retains the most important information. This compressed representation is the focus of this work and will be used for downstream analysis, clustering and classification.

Finally, the output of the autoencoder is passed through a set of individual linear layers to map the data back to its original dimension.

Given that the number of input features is 18, to construct the autoencoder it was decided that the first layer would have a smaller dimension to avoid expanding the feature space before the desired compression/bottleneck of the autoencoder. 

Table~\ref{tab:autoencoder_dim} shows the reconstruction error for different autoencoder architectures with varying layers dimensions. The results are given in the form of \textit{mean} ± \textit{std} of different trials. The notation used for the autoencoder dimensions represents the architecture of the neural network. The dimensions refer to the number of neurons or units in each layer of the network. The first and last dimensions correspond to the input and output layers, respectively, while the dimensions in between represent the hidden layers. For example, the representation \textit{[16, 8, 4, 8, 16]} indicates that autoencoder has five layers. The first layer has 16 neurons, which is the dimensionality of the input data. The subsequent layers progressively decrease in size, with 8 neurons in the second layer, and 4 neurons in the third layer (which acts as the bottleneck layer, where the input data is compressed into a lower-dimensional representation). Then, the layers increase back to 8 neurons in the fourth layer, and the last layer has 16 neurons, matching the dimensionality of the reconstructed output.

As can be seen, the best-performing architectures in terms of reconstruction error were \textit{[16, 4, 16]} and \textit{[16, 8, 4, 8, 16]}.

\begin{table}[!htb]
\renewcommand{\arraystretch}{1.3}
\caption{Comparison of autoencoder architectures and their corresponding reconstruction errors.}
\label{tab:autoencoder_dim}
\centering
\begin{tabular}{cc}
\hline
\textbf{Autoencoder}        &  \textbf{Reconstruction Error} \\
\textbf{Dimensions}        &  \textbf{[\textit{mean} ± \textit{std}]}  \\ \hline
{[}16, 8, 4, 2, 4, 8, 16{]} & 0.57 ± 0.13          \\
{[}16, 8, 4, 8, 16{]}       & 0.41 ± 0.08          \\
{[}16, 8, 2, 8, 16{]}       & 0.51 ± 0.13          \\
{[}16, 4, 2, 4, 16{]}       & 0.55 ± 0.11          \\
{[}16, 4, 16{]}             & \textbf{0.33 ± 0.08} \\
{[}16, 2, 16{]}             & 0.45 ± 0.10          \\ \hline
\end{tabular}
\end{table}

Based on common trends and best practices in autoencoder design, some conclusions can be retrieved from the results obtained. Firstly, the size of the latent space can have a significant impact on the performance of an autoencoder. A larger latent space allows the model to capture more complex patterns in the input data but also increases the risk of modelling noise in addition to the signal, i.e., overfitting. Contrarily, a smaller latent space may result in a simpler model, but may not be able to capture all of the important features in the data. The fact that the \textit{[16, 4, 16]} and \textit{[16, 8, 4, 8, 16]} architectures performed better suggests that the size of the latent space in these models strikes a good balance between complexity and simplicity, allowing the model to capture enough of the important features without overfitting.

Another factor to consider is the depth of the encoder and decoder networks. In general, deeper networks have more capacity to capture complex patterns but can be more difficult to train and may be more prone to overfitting. The \textit{[16, 8, 4, 8, 16]} architecture is deeper than the \textit{[16, 4, 16]} one, which may explain why it also performed well. However, the \textit{[16, 8, 4, 2, 4, 8, 16]} architecture achieved the highest reconstruction error, despite being even deeper. This suggests that there may be a point of diminishing returns where increasing the network depth no longer leads to significant improvements in performance. Thus, the chosen autoencoder architecture was the \textit{[16, 4, 16]} due to its performance.

\section{Latent Space Visualization and Clustering}

\begin{figure}[htb!]
    \centering
    \includegraphics[width=0.4\textwidth]{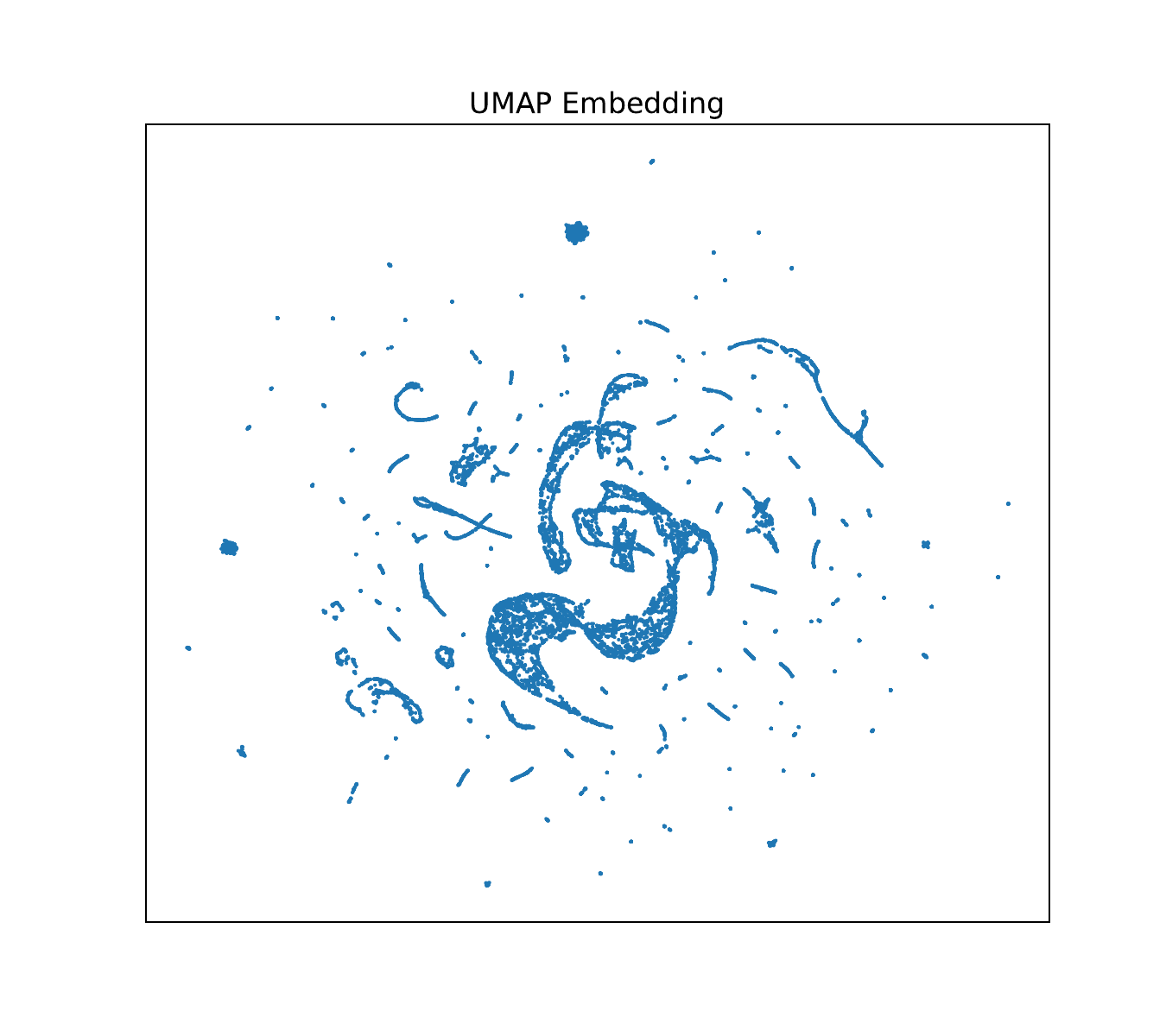}
    \caption{Two-dimensional UMAP visualisation of the latent space produced by the autoencoder.}
    \label{fig:umap_latentspace}
\end{figure}

Since the latent space has four dimensions, a dimensionality reduction technique is needed to visualize such high-dimensional data in a lower-dimensional space (Figure~\ref{fig:umap_latentspace}). To this end, the \gls{UMAP} algorithm was chosen due to its powerful advantages when compared to other methods~\cite{McInnes2018}.

The next step consisted in performing clustering on the latent space (Figure~\ref{fig:umap_latentspace}), with the aim of identifying different groups or patterns in the data.

Clustering is an unsupervised learning technique~\cite{Murphy2022} that partitions a dataset into groups of similar data points, called clusters, based on their similarity. In this study, we used the K-means clustering algorithm~\cite{Lloyd1957, MacQueen1967}, which is a widely used method~\cite{Murphy2022, Hastie2001} that aims to minimise the sum of squared distances between data points and their respective cluster centres.

To determine the optimal number of clusters, we computed the \gls{SSE} between each data point and its assigned cluster centre. The \gls{SSE} tends to decrease as the number of clusters increases because the data points can be divided into more specific and smaller groups. However, adding more clusters can also lead to overfitting, where the model is too complex and does not generalise well to new data. The elbow method is a commonly used heuristic to determine the optimal number of clusters based on the \gls{SSE}~\cite{Thorndike1953}. It involves plotting the \gls{SSE} as a function of the number of clusters and looking for an inflexion in the plot where the rate of decrease in \gls{SSE} slows down significantly. This bend indicates that the addition of more clusters beyond this point provides only marginal improvement in the \gls{SSE} and that the optimal number of clusters lies at or near this elbow point. 

\begin{figure}[htb!]
    \centering
    \includegraphics[width=0.45\textwidth]{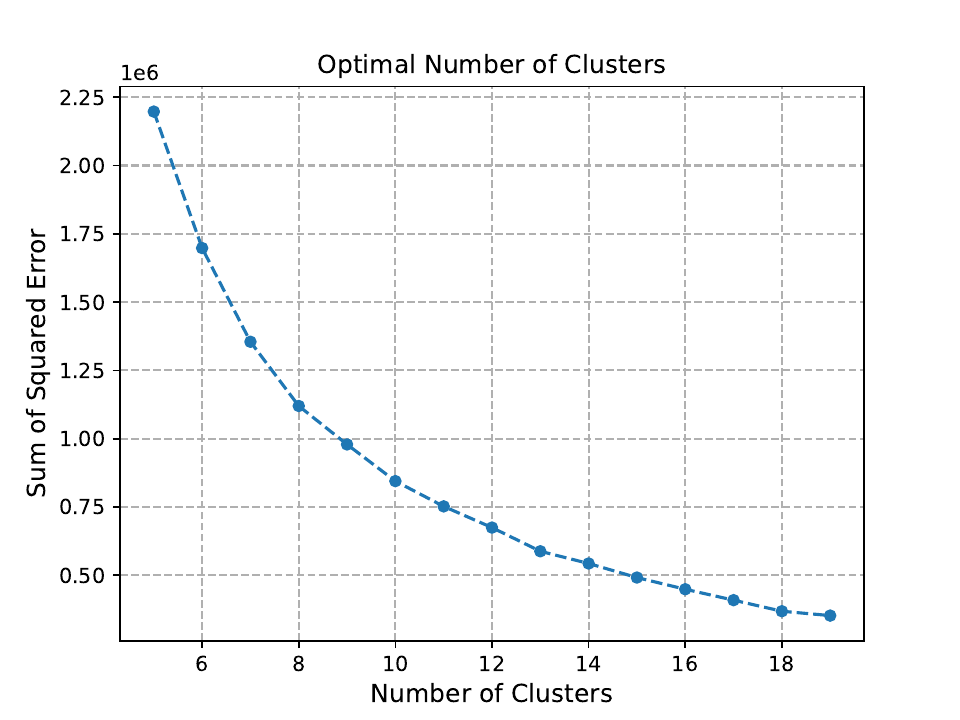}
    \caption{Sum of squared errors for different numbers of clusters in the latent space. The curve does not indicate a clear elbow point. However, it appears that the error decreases more slowly as the number of clusters increases beyond 8.}
    \label{fig:sse}
\end{figure}
 
Figure \ref{fig:sse} shows the \gls{SSE} for different numbers of clusters in the latent space. As can be seen, the curve can be ambiguous since it does not have a clear elbow point. However, it appears that the \gls{SSE} decreases more slowly as the number of clusters increases beyond 8. Besides, choosing a higher number of clusters may lead to overfitting or creating clusters that are too small to be meaningful. Therefore, we believe that 8 clusters is a reasonable choice for capturing the main patterns in the data.

After obtaining the optimal number of clusters, we applied K-means clustering on the latent space of the autoencoder to obtain the cluster labels for each data point.

\begin{figure}[htb!]
    \centering
    \includegraphics[width=0.45\textwidth]{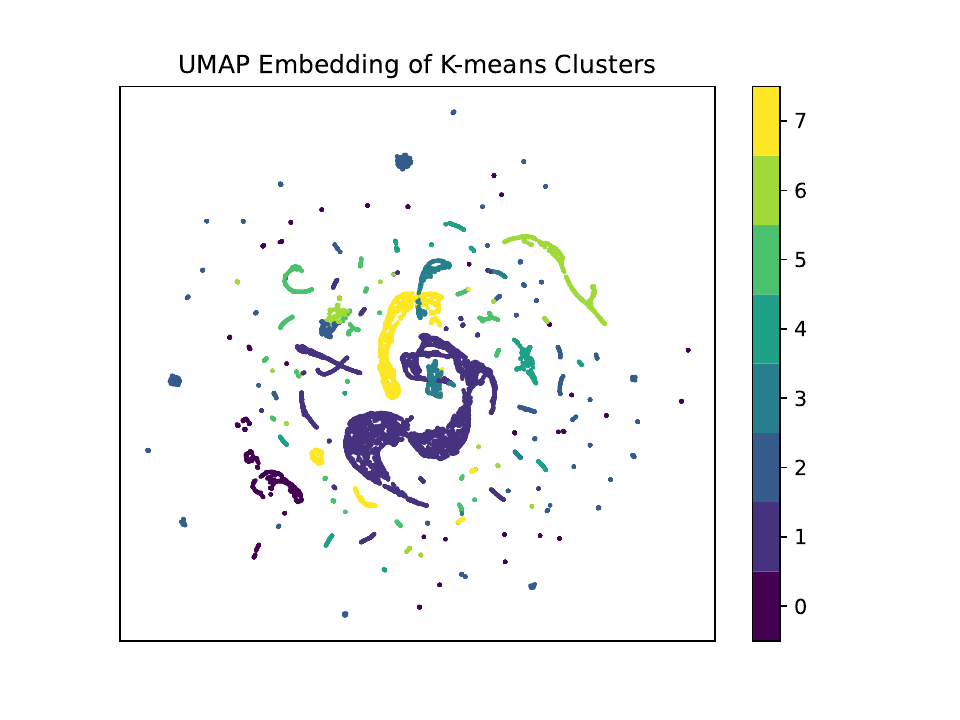}
    \caption{UMAP visualisation of the latent space coloured by the resulting K-means clusters. The resulting plot shows a clear separation between the different clusters, indicating that the K-means clustering was successful in identifying meaningful groups in the data. The colour bar indicates the legend of the different clusters.} 
    \label{fig:umap_kmeans}
\end{figure}

Figure~\ref{fig:umap_kmeans} shows the 2D projection of the latent space using the \gls{UMAP} algorithm,
with each data point coloured according to the cluster assignment obtained from K-means clustering. 
It can be seen that the clusters are relatively well-separated and identified, with minimal overlap between them, suggesting that the K-means algorithm was successful in identifying meaningful clusters in the latent space of the autoencoder.
The most evident cluster is the one presented in the centre in purple, i.e., cluster 1, indicating that there may be a strong underlying pattern in the data that distinguishes this group from the others. 
It can also be observed some areas where the clustering algorithm might have mislabeled some data points, as evidenced by the presence of small regions of a different colour within a larger cluster.

\section{Extracting Insights from the Clusters Using a Decision-Tree-Based Model} \label{sec:decision_tree}

After clustering the latent space and obtaining the cluster labels for each data point, the next step is to understand the characteristics of each cluster and extract meaningful rules that can be used to build the taxonomy. To achieve this, a classification task can be performed to predict the cluster label based on the input features. 

There are different supervised learning methods that can be used for classification tasks, such as neural networks, support vector machines and decision trees. Among these techniques, decision tree-based algorithms are one of the most commonly used methods in classification problems due to their interpretability achieved by the hierarchical placement of decisions~\cite{Murphy2022}. Furthermore, as we aim to extract meaningful rules to build the taxonomy, decision tree-based models are a good choice. Thus, it was decided to proceed with the LightGBM~\cite{Ke2017} model due to its capacity to handle missing data and categorical features, and its faster training speed when compared to other methods. The LightGBM model is an ensemble machine learning algorithm based on decision trees that uses boosting to improve the performance of the model.

To train the LightGBM model, we used the original input data as features and the cluster labels as target labels for the model. Note that in the global pipeline, the primary objective is to extract significant rules from the data, as opposed to estimating the generalization error of a predictive model. Therefore, the data was not split into training and testing sets, as the purpose is not to evaluate the performance of the model on unseen data. Instead, the entire dataset was used to train the LightGBM model and extract the rules from the data.

\begin{figure}[htb!]
    \centering
    \includegraphics[width=0.45\textwidth]{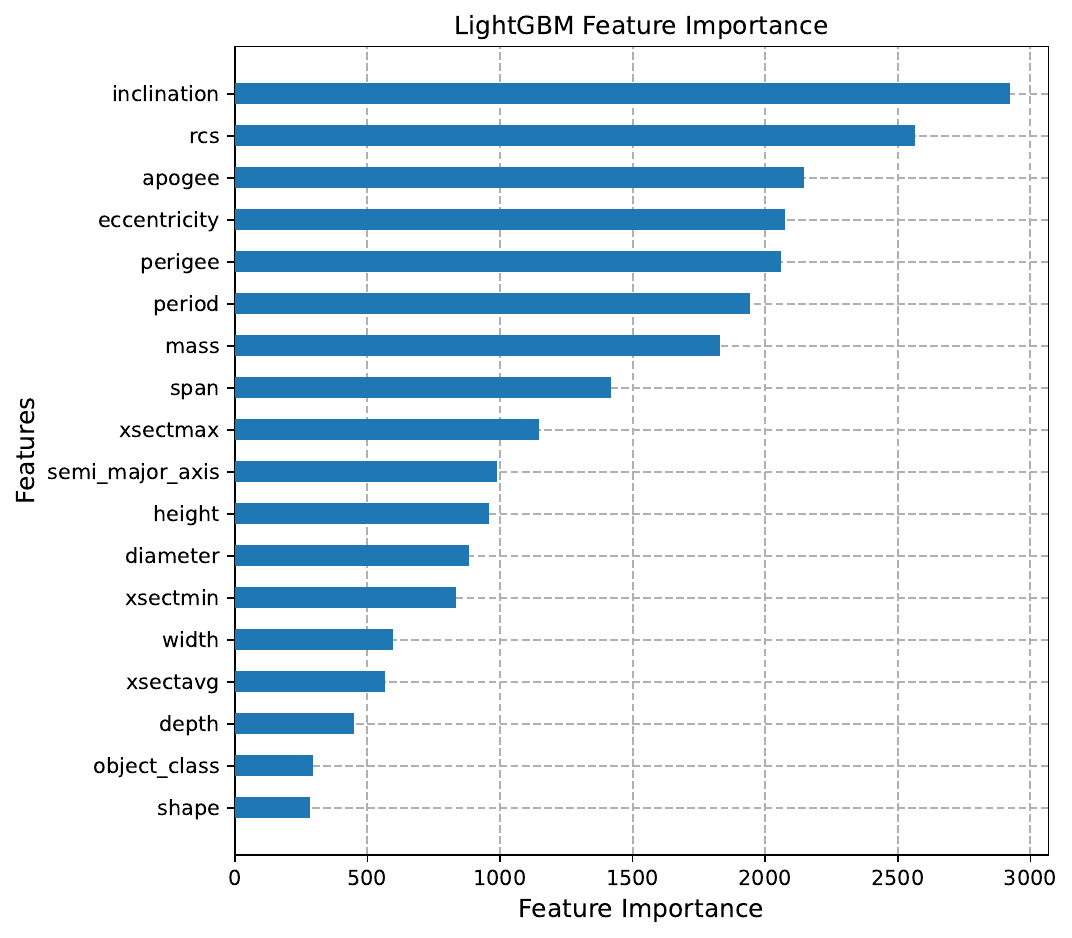}
    \caption{LightGBM feature importance. More relevant features are presented at the top.} 
    \label{fig:feature_importance}
\end{figure}

After training the model, we extracted the feature importance scores (Figure~\ref{fig:feature_importance}). Feature importance is computed based on the number of times a feature was used to split the data during training, and how much each split improved the model's performance.

From Figure~\ref{fig:feature_importance}, it can be concluded that the feature with the highest importance was the orbit inclination. This is one of the six Keplerian elements describing the shape and orientation of an orbit~\cite{Curtis2010}. It represents the angle between the reference plane and the orbital plane of the orbiting object.


From a domain perspective, the inclination is a crucial orbital parameter since it determines the coverage of the satellite and the accessibility of different regions of the Earth's surface. 
Satellites with an inclination close to 90 degrees, referred to as polar-orbiting satellites, pass over or near the Earth's poles on each orbit and cover the entire surface of the Earth over time (since the Earth rotates underneath the satellite as it orbits).
In contrast, satellites in \gls{LEO} with low inclinations, i.e., close to the equator, cover more limited areas of the Earth's surface but are better suited for specific applications such as remote sensing or communication.

The second feature with the greatest importance was the \gls{RCS}, which measures the object's ability to reflect radar signals in the direction of the radar receiver. The \gls{RCS} of a \gls{RSO} is influenced by a variety of factors, including the size, shape, and composition of the object. In general, larger \glspl{RSO} and those with complex shapes or sharp edges tend to have higher \gls{RCS} values.

The feature importance method of LightGBM calculates the contribution of each feature to the overall predictive accuracy of the model. It provides a ranking of the features based on their importance, but it does not tell us how each feature affects the prediction for a particular data point. Thus, a method typically used to address this problem is \gls{SHAP}~\cite{Lundberg2017}. Such a framework is used to explain the prediction of a certain instance by evaluating the contributions and importance of each feature to the results. 

The major advantage of using feature importance is that it provides a global ranking of the features that can be used to identify the most important features for the entire dataset. On the other hand, \gls{SHAP} values provide local explanations for individual data points, which can help to identify specific patterns or relationships between the features and the outcome. By using both methods, we can gain a better understanding of the overall patterns and relationships in the data. Thus, \gls{SHAP} values were also considered in the analysis. More concretely, we used TreeSHAP~\cite{Lundberg2020}, a variant of \gls{SHAP} for tree-based machine learning models such as the LightGBM.

\begin{figure}[htb!]
    \centering
    \includegraphics[width=0.45\textwidth]{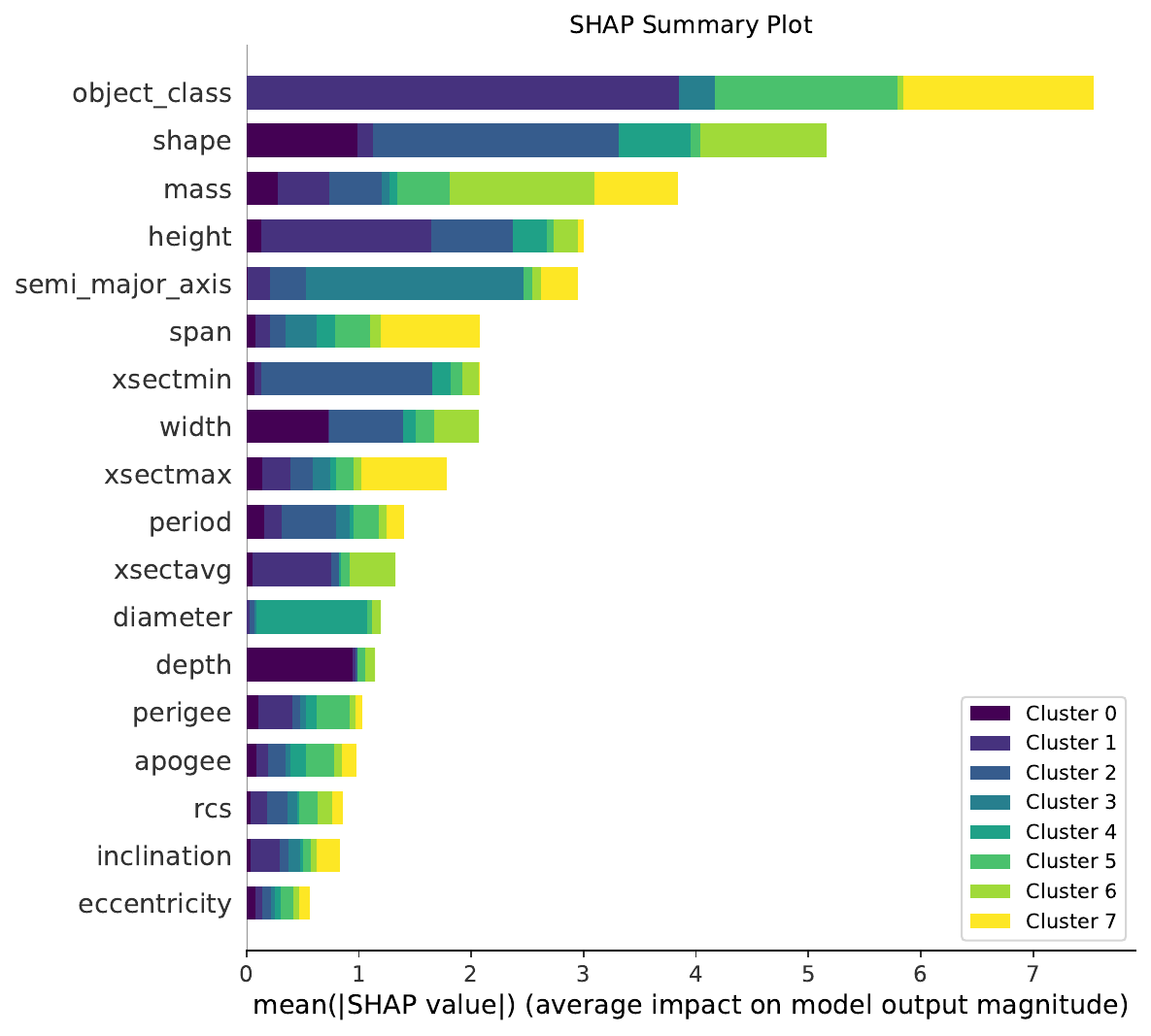}
    \caption{Summary plot of SHAP values for the LightGBM model. The importance is defined as the mean absolute value of the SHAP values for each feature on all clusters. The x-axis represents the magnitude of the Shapley values and the position along the y-axis is determined by the feature. More relevant features are at the top. } 
    \label{fig:shap}
\end{figure}

Figure~\ref{fig:shap} shows the \gls{SHAP} summary plot, which combines feature effects and feature importance, giving a post hoc explainability of the model. Note that the impact of a feature on the clusters is stacked, as the problem is a multi-classification task. It can be seen that some features have a significant effect on multiple clusters (e.g., shape and mass), while others only affect a few clusters (e.g., diameter and depth). The feature with the greatest impact is the object class (retrieved from \gls{DISCOS}). This is an indicator that the different types of objects have unique characteristics and behaviours that affect their distribution in the \gls{LEO} environment. For instance, it is known that rocket bodies tend to be larger and heavier than other debris types, and thus more likely to experience different forces and interactions with the Earth's atmosphere and gravitational field. Payloads, on the other hand, may have different materials or configurations that affect their fragmentation and decay rates.

\begin{figure}[htb!]
    \centering
    \includegraphics[width=0.45\textwidth]{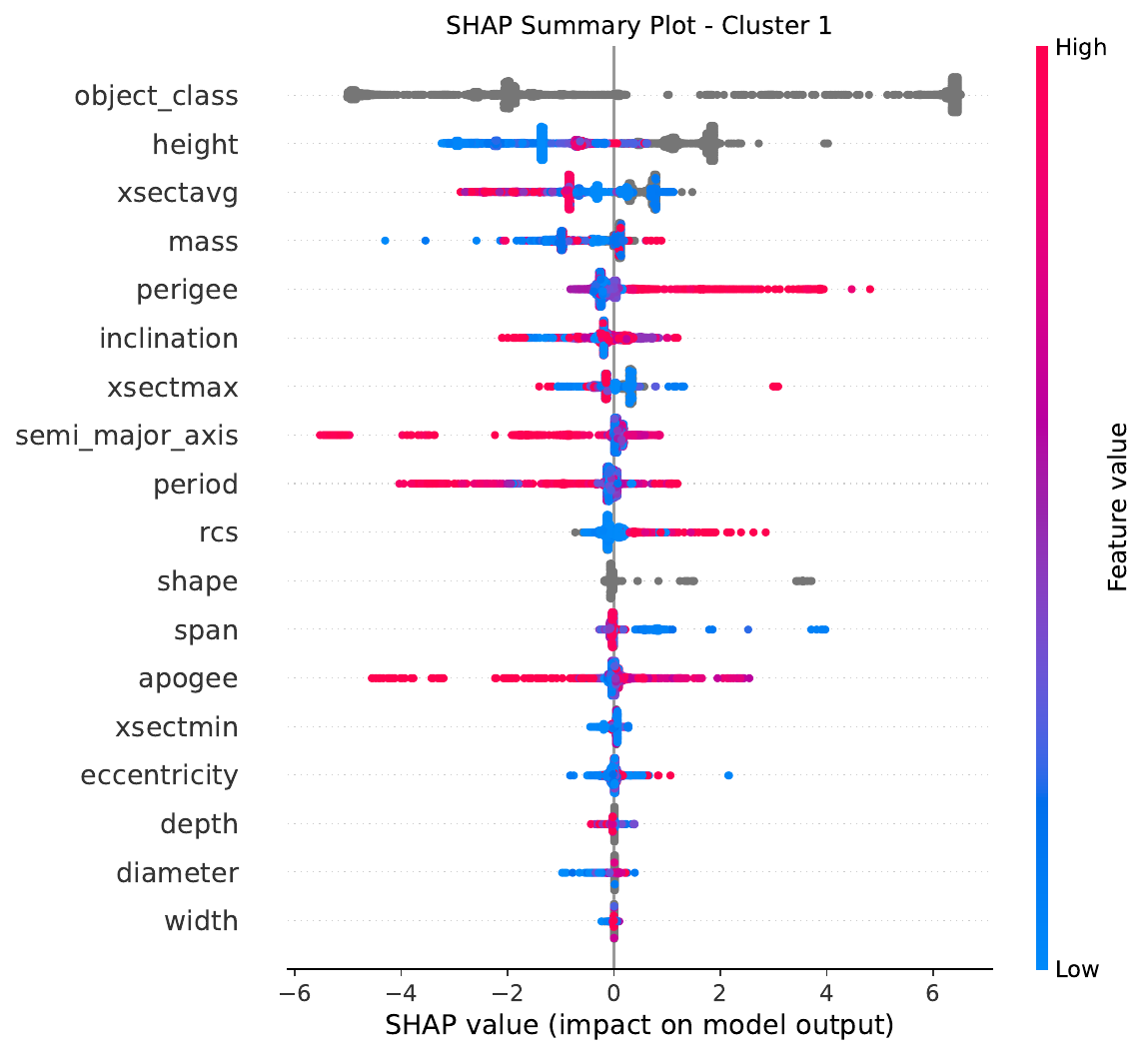}
    \caption{SHAP summary plot of cluster 1. The colour map represents the feature value on a scale from low to high. For each feature, there are some jittered points along the y-axis giving a representation of overlapping points. Categorical features (i.e., object class and shape) and missing values from the real features are presented in grey.} 
    \label{fig:shap_cluster1}
\end{figure}

To get a better understanding of the impact of the features on a specific cluster, we can plot the \gls{SHAP} summary plot for that given cluster. As seen in Figure~\ref{fig:umap_kmeans}, the cluster with more samples is cluster 1. Thus, it was decided to further analyse such a cluster. From Figure~\ref{fig:shap_cluster1} it can be confirmed that indeed, the feature with the greatest importance is the object class. Note that since this variable is categorical it is not possible to distinguish which classes are more relevant to the model output. The second feature with the greatest importance is the object's height (retrieved from \gls{DISCOS}). Interestingly, samples with missing values have positive \gls{SHAP} values. This means that the missing height values are more likely to contribute to a positive prediction, i.e., to belong to cluster 1. 

\begin{figure}[htb!]
    \centering
    \includegraphics[width=0.45\textwidth]{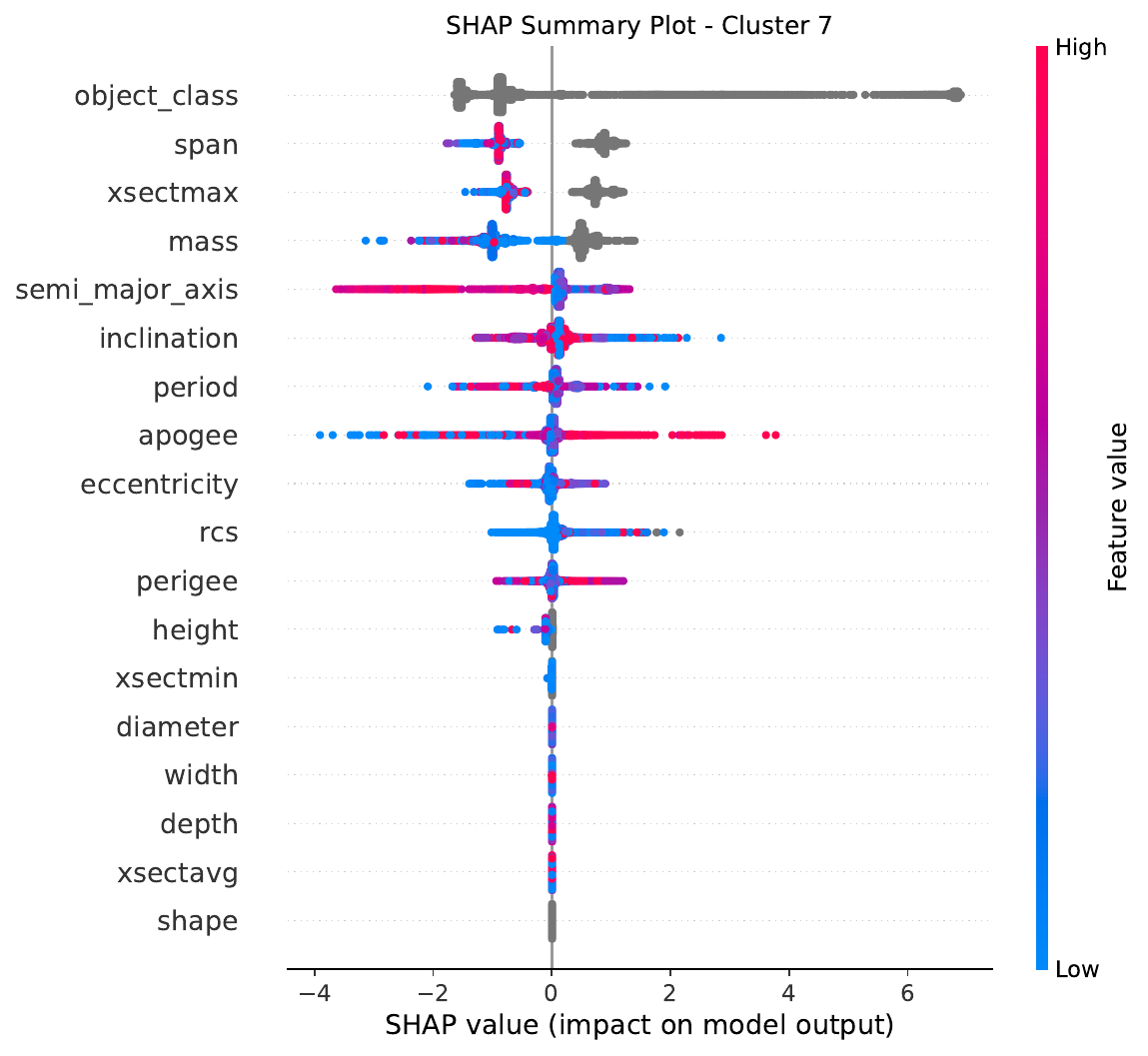}
    \caption{SHAP summary plot of cluster 7. Categorical features (i.e., object class and shape) and missing values from the real features are presented in grey. Note that the features with the highest features importance are presented at the top.} 
    \label{fig:shap_cluster7}
\end{figure}

Another cluster that stands out in Figure~\ref{fig:umap_kmeans} is cluster 7 (in yellow). From Figure~\ref{fig:shap} it can be seen that the features that impact the most this cluster are the object class, mass and span. To gain further insight into this cluster, we can examine the \gls{SHAP} summary plot for the objects in this cluster (Figure \ref{fig:shap_cluster7}).

From this plot, we can confirm that the most important feature for distinguishing objects in this cluster is the object class, their span, followed by their \gls{RCS} and their mass. This indicates that the size and shape of these objects are important factors in determining their membership in the cluster. As seen in cluster 1, the presence of missing values in the dataset can provide valuable information about the clustering of \glspl{RSO}. This highlights the importance of not assuming default values in our study, and of considering the presence of missing values as additional information.

\section{Taxonomy}

The LightGBM feature importance analysis highlighted the significance of the inclination, \gls{RCS}, and apogee in determining the behaviour of \glspl{RSO} in the \gls{LEO}. On the other hand, the \gls{SHAP} summary plot emphasized the importance of object class, shape, and mass in differentiating between the clusters. These findings highlight the importance of considering both the main characteristics of the objects, as well as their localisation in the orbit to create a comprehensive taxonomy. Therefore, we propose two hierarchical taxonomies, to provide a more complete understanding of the space debris population in \gls{LEO}. The first category focuses on the main characteristics of the objects, including their class, shape, and mass. The second category handles the localisation of the objects in orbit, including parameters such as orbital inclination. It is important to note that an object belongs to a leaf of both taxonomies at the same time, as both its unique characteristics and its orbit parameters are relevant for understanding its behaviour and potential risks. For a given object, the taxonomy is defined by a tree which is obtained by the product of the decision on each level.

\subsection{Object Characteristics Taxonomy}

\begin{figure*}[t!]
    \centering
    \includegraphics[width=\textwidth]{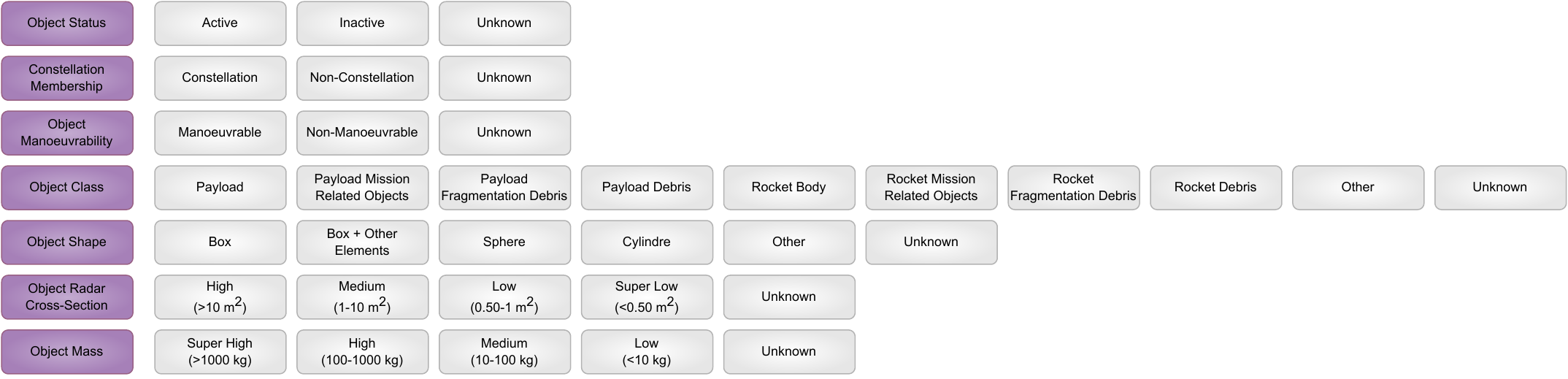}
    \caption{Object characteristics taxonomy. The levels should be interpreted from the top to the bottom, i.e., the ``Object Status'' represents the first level, and the ``Object Mass'' defines the last one, representing the leaves of the decision tree. Note the presence of the ``Unknown" category in all the levels.} 
    \label{fig:taxonomy1}
\end{figure*}

The proposed hierarchy for the object characteristics taxonomy provides a comprehensive view of the key characteristics of space objects in \gls{LEO} (Figure~\ref{fig:taxonomy1}).

The first level focuses on the current status of the object, i.e., if the object is active, inactive or if the current status is unknown. Despite not being one of the parameters used to train our model, it is an important factor for understanding the behaviour and potential risks of a given object.
The second level assesses whether or not the object is part of a constellation. This information is relevant for understanding the behaviour of the objects in orbit.
The third level takes into account the manoeuvrability of the object, which is relevant for predicting its future trajectory and for assessing the potential risks of collision. Note that, similarly to the object status, levels two and three features were not available to train the model. We recommend that future work introduces such variables in the dataset, to explore the impact of these values in the clustering and model explainability.
The fourth level focuses on the object class, which is a piece of valuable information for understanding the object's function and its potential for being a source of debris. The different categories correspond to the ones used in our model, retrieved from \gls{DISCOS} dataset. It was decided to keep the same categories as they are well-established and accepted by industry practitioners~\cite{Merz2017}.
The fifth level considers the object's shape, which provides insights into the potential for collision with other objects, as well as the object's stability. The different categories were also retrieved from \gls{DISCOS} dataset, but due to the high cardinality of these features only the most common shapes were considered in the taxonomy. Future work could explore the usage of more categories at this level and consider the known emerging satellite shapes, such as flat-shaped satellites. 
The sixth level is the \gls{RCS}. This is an important factor for space situational awareness, as objects with higher \gls{RCS} can be more easily tracked and monitored. Note that while varying with material composition, overall shape and structure, the \gls{RCS} of an irregular object is also dependent on its spatial orientation. However, typically the main contributions come from the object's characteristics. To define the different ranges of values in our taxonomy, we have used the distribution of this variable in our dataset. It is important to highlight that the datasets used in this work do not account for the usage of retroreflectors. Thus, future work should analyse such variable carefully. 
The last level accounts for the object mass, which is an important factor in determining the potential for damage upon collision, as well as manoeuvres planning. For this level, the thresholds were defined based on the values used on the \gls{ESA}'s annual report~\cite{ESA2022}.

\subsection{Object Orbit Localisation Taxonomy}

As already mentioned, one of the features with the greatest importance on the LightGBM model was the apogee. However, such a feature is typically associated with the perigee to provide more valuable information. Thus, the first level of this taxonomy is based on the semi-major axis of the object's orbit, which is the sum of the perigee and apogee divided by two. It provides a general idea of the altitude range at which the object is orbiting. This information is relevant for predicting the object's behaviour over time and assessing the effects of atmospheric drag on the object's trajectory. Additionally, the semi-major axis can also provide insights into the type of mission that the satellite may have, as certain orbits are better suited for specific applications.

\begin{figure}[htb!]
    \centering
    \includegraphics[width=0.45\textwidth]{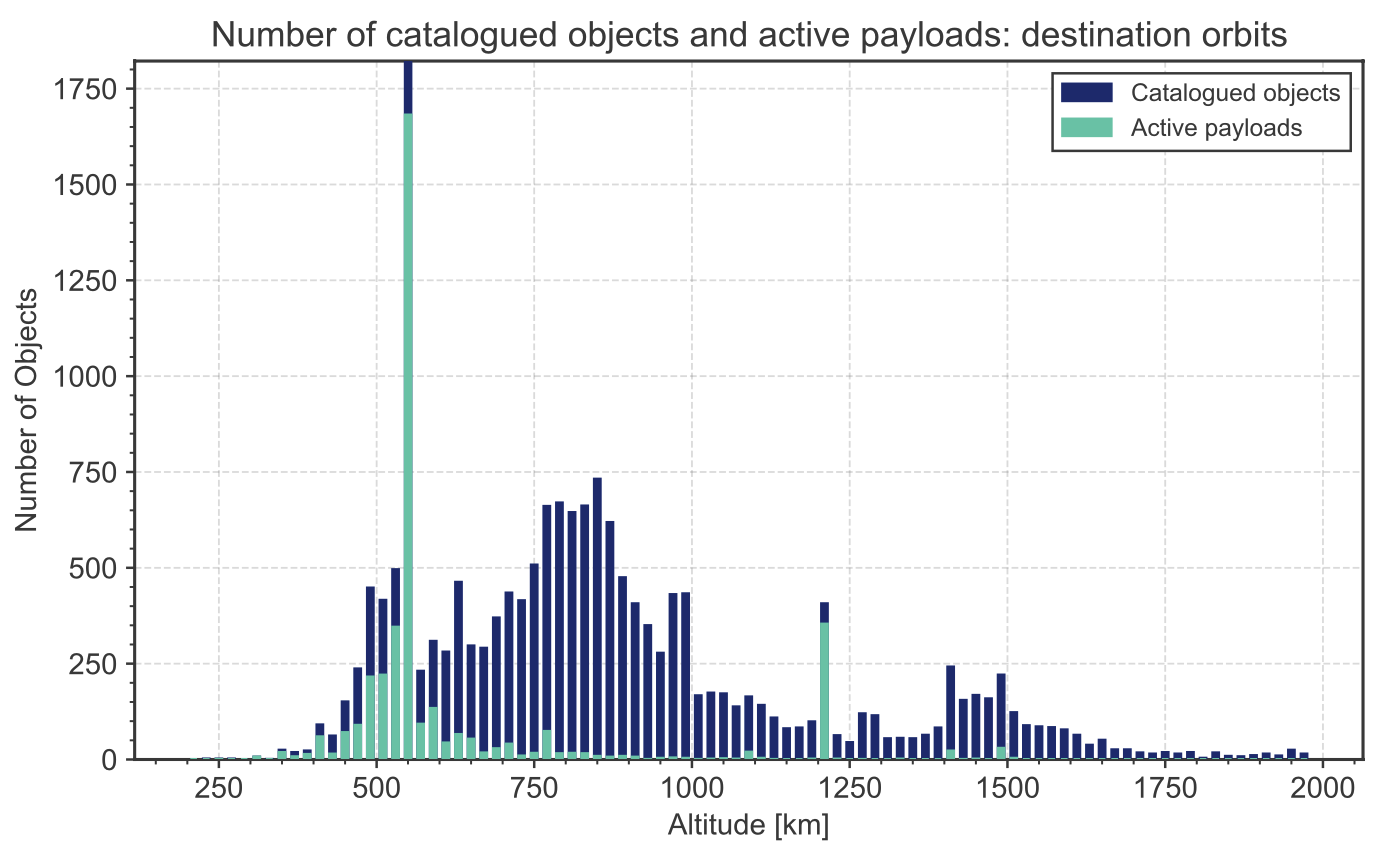}
    \caption{Distribution of the altitude of the catalogued objects and active payloads as reported by ESA~\cite{ESA2022}. It can be seen that there are two primary ``active payload'' regions around 500 and 1,250 km.} 
    \label{fig:esa_altitude_distribution}
\end{figure}

From Figure~\ref{fig:esa_altitude_distribution}, it can be seen that there are two primary ``active payload'' regions around 500 and 1,250 km. Moreover, a major region of other ``catalogued objects'' is also present from 600 to 1,000 km. Thus, these thresholds were the ones used to build the taxonomy.

\begin{figure}[htb!]
    \centering
    \includegraphics[width=0.45\textwidth]{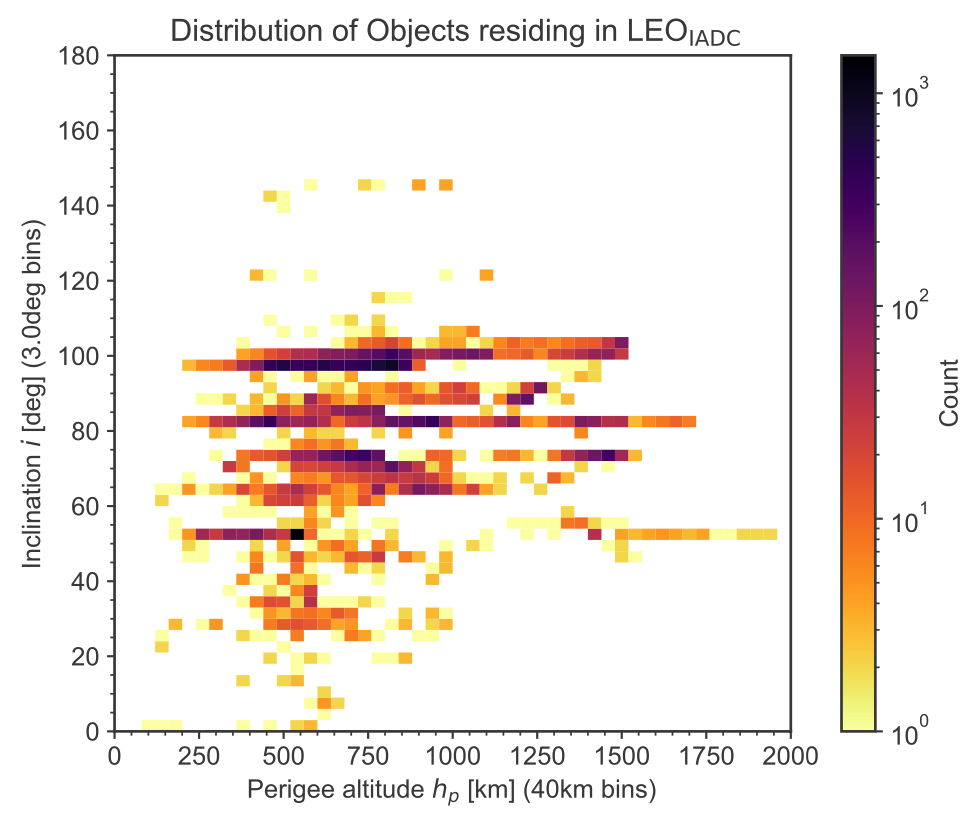}
    \caption{Distribution of inclination and perigee altitude as reported by ESA~\cite{ESA2022}. There are inclination intervals where the high distribution of objects is evident (dark purple squares).} 
    \label{fig:esa_inclination_distribution}
\end{figure}

\begin{figure*}[htb!]
    \centering
    \includegraphics[width=\textwidth]{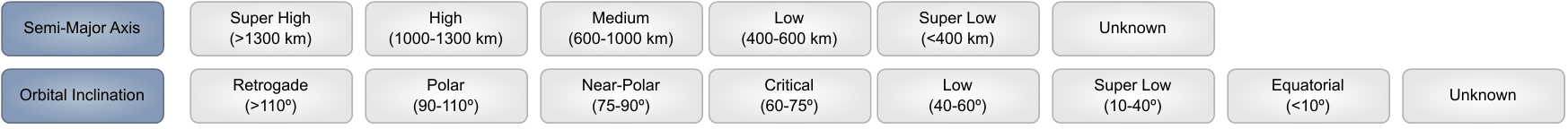}
    \caption{Object orbit localisation taxonomy. The levels should be interpreted from the top to the bottom, i.e., the ``Semi-Major Axis'' represents the first level, and the ``Orbit Inclination'' defines the last one. Note that an object simultaneously belongs to a leaf of both taxonomy trees, i.e., Figures \ref{fig:taxonomy1} and \ref{fig:taxonomy2}.} 
    \label{fig:taxonomy2}
\end{figure*}

The second level of this taxonomy is the orbital inclination of the object. As mentioned in Section~\ref{sec:decision_tree}, the inclination measures the angle between the reference plane and the object's orbital plane. Furthermore, the inclination can also provide insights into the object's purpose, as certain missions require specific inclinations for optimal performance. The thresholds used to bin the orbital inclination were based on the values from Figure~\ref{fig:esa_inclination_distribution}. As can be seen, there are clear inclination intervals where the high distribution of objects is evident. Based on such values, the different categories for our taxonomy were defined (Figure~\ref{fig:taxonomy2}).

\section{Conclusion}
We proposed a novel taxonomy for \glspl{RSO} in \gls{LEO}, derived from a comprehensive analysis of a large dataset of catalogued objects. The combination of real-valued and categorical variables in our approach enables a comprehensive assessment of the intrinsic characteristics of \glspl{RSO}, including their object class, shape, and more. Moreover, our approach to handling missing values is a significant advancement in the field. Rather than relying on imputation or assuming default values, we recognise the value of missing data as an informative feature itself. By considering missing values as additional information, we ensure that our taxonomy reflects the true nature of the objects and avoids potential biases that could arise from imputation.

We presented a \gls{DL}-based model that uses an autoencoder architecture to reduce the features representing the characteristics of the \glspl{RSO}. The autoencoder generates a lower dimensional space representation that is then explored using techniques such as \gls{UMAP} to identify fundamental clusters of \glspl{RSO} based on their unique characteristics. This approach captures the complex and non-linear relationships between the features and the \glspl{RSO}’ classes identified. 

A future line of work could expand this study to include additional features and parameters. For example, incorporating time-dependent variables can provide valuable insights into the dynamics and evolution of objects over time. Besides, the integration of information regarding the level of automation of the \glspl{RSO} would yield a more comprehensive understanding of their operational characteristics, and highlight the trends in automation adoption and its implications for space traffic management.

Our proposed taxonomy and model offer a significant contribution to the ongoing efforts to mitigate the overall risks posed by the increasing number of \glspl{RSO} in orbit.

\section*{Acknowledgements}
This research was carried out under Project “Artificial Intelligence Fights Space Debris” Nº C626449889-0046305 co-funded by Recovery and Resilience Plan and NextGeneration EU Funds (www.recuperarportugal.gov.pt), and by NOVA LINCS (UIDB/04516/2020) with the financial support of FCT.IP.

%

\bibliographystyle{unsrtnat}

%



\end{document}